\documentclass[runningheads]{llncs}

\usepackage{eccv}

\usepackage{eccvabbrv}

\usepackage{graphicx}
\usepackage{booktabs}
\usepackage{multirow}
\usepackage[accsupp]{axessibility}  

\usepackage{xcolor} 

\usepackage{hyperref}
\usepackage{wrapfig}

\usepackage{orcidlink}

\begin{document}

\title{CadVLM: Bridging Language and Vision in the Generation of Parametric CAD Sketches} 

\titlerunning{CadVLM}

\author{Sifan Wu\inst{1}\orcidlink{0000−0001−5752−0929} \and
Amir Hosein Khasahmadi\inst{2} \and
Mor Katz\inst{2} \and
Pradeep Kumar Jayaraman\inst{2} \and
Yewen Pu\inst{2} \and
Karl Willis\inst{2} \and
Bang Liu\inst{1}\orcidlink{0000-0002-9483-8984}} 

\authorrunning{S.~Wu et al.}

\institute{Université de Montréal \& MILA, Montreal, Canda \and
Autodesk AI Lab, Toronto, Canada} 
\maketitle

\begin{figure*}
    \includegraphics[width=0.98\textwidth]{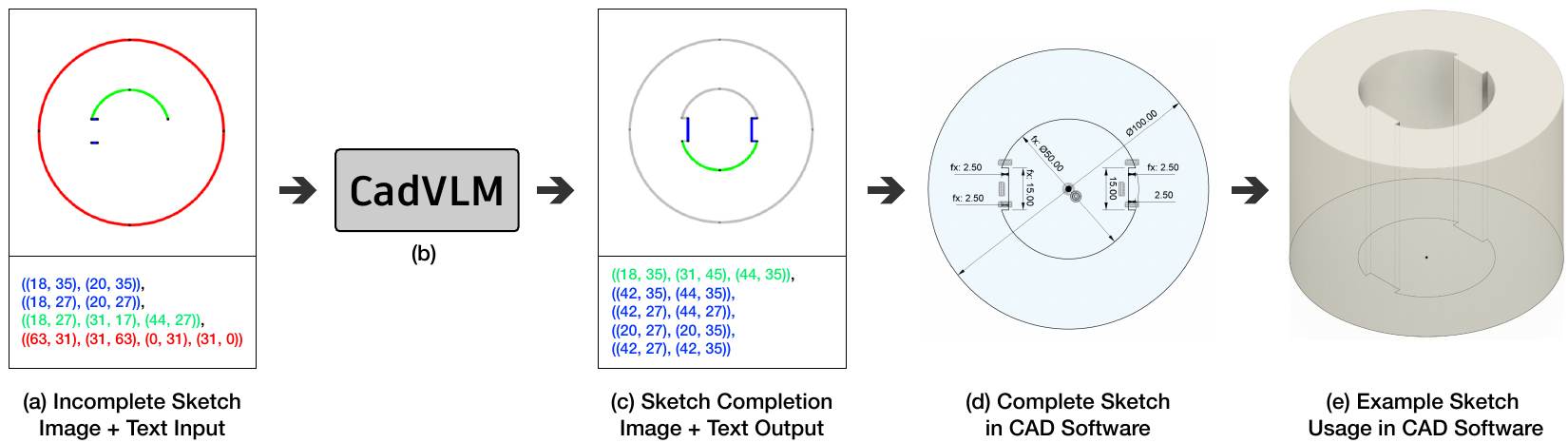}
    \caption{For the CAD autocompletion task, our multi-modal \textbf{CadVLM} model (b) receives partial CAD entities as both image and text input (a) and generates the remaining sketch entities as output (c). The complete sketch, with optional predicted constraints, can then be used in CAD software (d) to form 3D shapes (e).More description about the primitive values in sketch text are in the Appendix.}
    \label{fig:intro}
\end{figure*}
\begin{abstract}
Parametric Computer-Aided Design (CAD) is central to contemporary mechanical design. However, it encounters challenges in achieving precise parametric sketch modeling and lacks practical evaluation metrics suitable for mechanical design.
We harness the capabilities of pre-trained foundation models, renowned for their successes in natural language processing and computer vision, to develop generative models specifically for CAD. These models are adept at understanding complex geometries and design reasoning, a crucial advancement in CAD technology. In this paper, we propose CadVLM, an end-to-end vision language model for CAD generation. Our approach involves adapting pre-trained foundation models to manipulate engineering sketches effectively, integrating both sketch primitive sequences and sketch images. Extensive experiments demonstrate superior performance on multiple CAD sketch generation tasks such as CAD autocompletion, CAD autoconstraint, and image conditional generation. To our knowledge, this is the first instance of a multimodal Large Language Model (LLM) being successfully applied to parametric CAD generation, representing a pioneering step in the field of computer-aided mechanical design. 
\end{abstract}


\section{Introduction}
\label{sec:intro}

Parametric Computer-Aided Design (CAD) is crucial in mechanical engineering, playing a key role in developing modern objects such as lamps, cars, and buildings. Its importance lies in its ability to accurately depict complex designs, essential for the integrity and functionality of engineered products interacting with humans and the environment. CAD sketches, consisting of basic 2D shapes, are foundational for parametric CAD modeling~\cite{camba2016parametric}. As illustrated in Fig~\ref{fig:intro},  these sketches comprise primitive geometric elements (e.g., lines, arcs, points) defined by various relationships (e.g., coincident, parallel, tangent). However, the required precision for 3D construction in parametric CAD can be both challenging and time-consuming~\cite{lindsay2018identifying}. Consequently, there is a significant push to improve CAD generation to simplify and accelerate the manual design process. Leveraging deep learning to identify patterns in CAD sketches could automate the completion of geometry and constraints in engineering drawings~\cite{seff2021vitruvion, seff2020sketchgraphs, willis2021engineering}.

Integrating deep learning into CAD offers a significant opportunity to transform the field. 
Recent research~\cite{seff2020sketchgraphs,seff2021vitruvion, li2023secad, jones2023self, wu2021deepcad, para2021sketchgen, xu2022skexgen} has concentrated on the generative modeling of engineering sketches, particularly focusing on Transformer-based architectures ~\cite{vaswani2017attention}. These approaches often utilize graph structures~\cite{seff2020sketchgraphs} or attention mechanisms~\cite{seff2021vitruvion} to process sketch information. However, the effectiveness of these methods is limited by the availability of data and a lack of commonsense understanding, which restricts their capacity to perform complex geometric reasoning in engineering sketches. Additionally, rendered sketch images (such as in Fig. \ref{fig:intro}(a) and \ref{fig:samples}) have great potential to represent geometry patterns from image modality, which could be complementary for parametric sketches language. While none of current CAD generative models take sketch image into consideration.
Our research presents a novel Multimodal Transformer-based Generative model for CAD generation tasks, pioneering in its use of both language and image for sketch modeling. By merging these modalities, we aim to increase the accuracy of CAD generation, setting a new benchmark by simultaneously modeling text-style sketch parametric and image-style sketch renderings.

Large Language Models (LLMs) have gained widespread success for their powerful commonsense reasoning and problem-solving abilities in a variety of domains, 
such as natural language processing, mathematics, and medical sciences ~\cite{singhal2022large,azerbayev2023llemma, luo2023wizardmath, singhal2023towards,floridi2020gpt, chen2022pali,zha2023tablegpt}. The capabilities of LLMs align well with engineering sketches, which can be understood as sequences representing points and connections. Commonsense knowledge in pre-trained LLM can also offer valuable hints for sketch patterns. Furthermore, combining sketch images with textual descriptions can greatly improve the performance of language models in CAD sketch generative models. 
In the realm of parametric CAD design, which requires a thorough understanding of complex interactions(e.g. parallel, coincident, symmetry), the advantages of LLMs become even more pronounced.


In this work, we leverage the capabilities of pre-trained LLMs to develop a generative model tailored for parametric CAD, addressing the limitations of the existing methods. We introduce \textbf{CadVLM}, a CAD Vision Language Model, featuring a multimodal encoder-decoder framework. CadVLM stands out by processing engineering sketches both linguistically, through token sequences representing sketch geometry and constraints, and visually, through the use of rendered sketch images. To the best of our knowledge, CadVLM is the first generative model to utilize pre-trained vision and language models for CAD sketches, which is capable of addressing different CAD representation learning tasks such as CAD autocompletion, CAD autoconstraint, and image conditional generation tasks.
Our contributions are threefold:
\begin{itemize}
\item We propose CadVLM, a novel CAD Vision-Language Model, for simultaneously modeling engineering sketches as both language and images. As we know, we are the first to combine visual and textual data for CAD generative models.
\item We introduce three novel evaluation metrics—Entity Accuracy, Sketch Accuracy, and CAD F1 score—to quantitatively assess the quality of generated CAD sketches.
\item 
CadVLM demonstrates superior performance on the SketchGraphs dataset~\cite{seff2020sketchgraphs} in both CAD autocompletion and autoconstraint tasks.
\end{itemize}

\section{Related Work}
\label{sec:relate_work}
\textbf{Vision Language Pre-Training.}
The great success of large language models (LLMs) such as BERT~\cite{kenton2019bert} and GPT~\cite{brown2020language} has drawn significant attention within the field of natural language processing and code generation domain. 
CodeT5+~\cite{wang2023codet5+} is a family of open-source encoder-decoder LLMs for code tasks, ranging from 220 million to 16 Billion parameters. 
Beyond single modality LLMs, multimodal foundation models have also exhibited enhanced performance in various vision-and-language tasks~\cite{radford2021learning, chen2024camml, li2022blip}. Depending on downstream tasks, 
Various pre-training objectives have also been proposed like image-text contrastive learning~\cite{radford2021learning}, image-text matching~\cite{li2022blip}, and (masked) language modeling~\cite{li2022blip}.
Most recently,  developing domain-specific multimodality LLMs has gained momentum such as MedBLIP~\cite{chen2023medblip} and TableGPT~\cite{zha2023tablegpt}.

\textbf{CAD Sketch Generation}
2D engineering sketches are fundamental in CAD for designing and manufacturing mechanical parts. 
The recent availability of extensive engineering sketch datasets has facilitated the development of generative models that enhance traditional CAD workflows. The SketchGraphs~\cite{seff2020sketchgraphs} dataset, featuring 15 million parametric CAD sketches with detailed geometric constraint graphs, showcases primitive interrelations. The paper also introduces a baseline model using autoregressive message-passing networks for constraint graph generation, but it lacks primitive coordinate output, relying solely on constraint graphs and the Onshape solver for sketch configuration, thus limiting its utility in CAD autocompletion tasks.
Ganin et al. \cite{ganin2021computeraided} proposed another large-scale CAD dataset containing over 4.7M parametric sketches from the Onshape public repository. However, limited by sketch format, this dataset has not been widely researched in the realm of CAD generative models.

Another line of work for generative modeling of CAD sketches is based on Transformer~\cite{willis2021engineering,wu2021deepcad, seff2021vitruvion, xu2023hierarchical}. Vitruvion~\cite{seff2021vitruvion} models both primitives and constraints, but its training on Transformer does not inherently include geometric knowledge. Furthermore, Vitruvion's major limitation lies in its lack of quantitative comparison in terms of CAD generation quality.
While CurveGen~\cite{willis2021engineering} tackles the problem of engineering sketch generation without considering sketch constraints like us, it struggles to generalize to few-shot samples. Both Vitruvion and CurveGen face limitations in their model performance, making them challenging to deploy in practical CAD engineering applications. Recently, there has been some research about CAD construction sequences generation~\cite{xu2022skexgen,xu2023hierarchical}. However, these works cannot model sketch primitives to autocomplete parametric CAD sketches. 
We consider Vitruvion, Deepcad, and SketchGraphs as the baselines for CAD Autocompletion task and CAD Autoconstraint task since they are are current state-of-the-art models using transformer architectures.
Different from existing work, our method utilizes sketch image modality and commonsense knowledge in pre-trained foundation models, providing both sketch sequence textual and image information for sketch generation.


\section{Background}


\textbf{Engineering Sketch Definition.} In the field of parametric CAD, engineering sketches are fundamental, serving as detailed blueprints for designing a range of objects from small components to large structures. These sketches are essential for converting conceptual ideas into precise, manufacturable designs. To ensure a standardized representation, sketches of varying dimensions, spanning from millimeters to meters, are uniformly positioned within a bounding box of 1-meter width at the origin. Each sketch consists of two key sequences: a sequence of primitives, $\mathcal{S}$, and a sequence of constraints, $\mathcal{C}$. 

We adopt the normalization and quantization methods for primitive parameters as outlined by~\cite{seff2021vitruvion}. A sketch primitive is expressed as $\mathcal{S} = (e_1, e_2, ..., e_m),$ where $m$ represents the number of entities in the sketch. Each entity $e_i$ is characterized by positional parameters $(p_1,..., p_k)$, where $p$ represents the normalized coordinates of points that constitute the entity. Sketches typically include three types of entities—lines, arcs, and circles. 
As illustrated in Figure~\ref{fig:intro}, all sketches are centrally positioned and normalized within a range of [1,64]. Lines, shown in blue, are depicted using two points: start and end point coordinates. Curves, shown in green, are characterized by three points: the start, the end, and a mid-point on the curve. Circles, illustrated in red, are portrayed with four uniformly distributed points on the circle's circumference. This format allows for the interpretation of each sketch as a sequence of tokens. Constraints are similarly represented as primitives in $\mathcal{C} = (c_1, ..., c_n),$ with each constraint entity $c$ being defined by its type and the reference indices of parameters to which the primitives must conform.

\label{sec:background}
\begin{figure*}[t]
    \centering
    \includegraphics[scale=0.37]{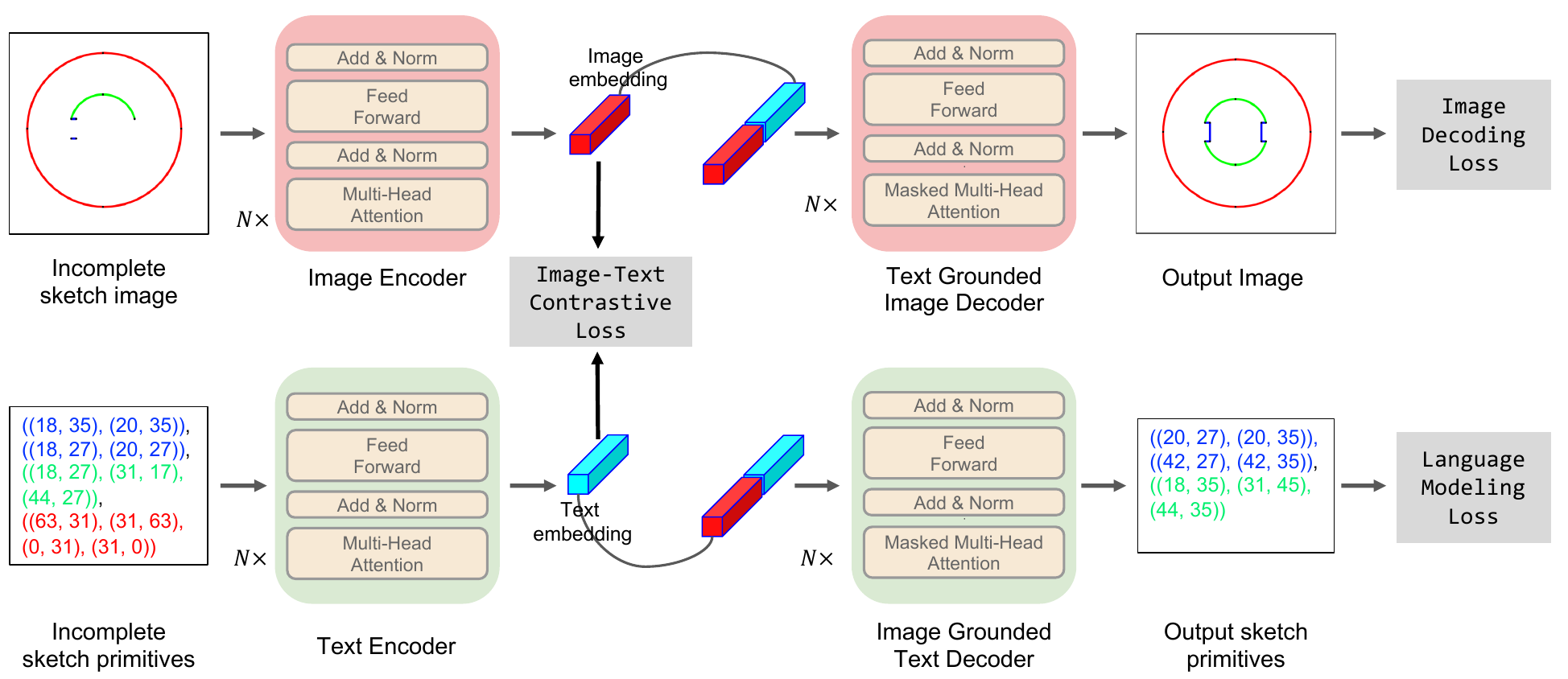}
    \caption{\textbf{CadVLM model architecture and training objectives} for CAD autoconstraint. We propose a unified vision-language model capable of outputting both autocompleted sketch images and sketch primitives. The inputs are paired with incomplete sketch primitives and images, which are encoded by pre-trained text encoder and image encoder correspondingly. Aligned by image-text contrastive(ITC) loss, the concatenation of image embedding and text embedding then will be input to the text-grounded image decoder and image-grounded text decoder. The image decoder is trained with image decoding loss between autocompleted sketch images and ground-truth sketch images. The text decoder is trained with a language modeling(LM) loss to generate autocompleted primitive.}
    \label{fig:architecture}
\end{figure*}

\textbf{CAD AutoCompletion Task.} In addressing the existing CAD AutoCompletion task, which aims to automate routine CAD design procedures, we propose a new autoregressive generative model. This task involves completing a partially given sketch to form a fully realized design. The mathematical formulation of the task is:
\begin{equation}
\setlength{\abovedisplayskip}{2pt}
\setlength{\belowdisplayskip}{2pt}
    \mathcal{L}(\Theta) = - \log p_{\Theta}(\mathcal{S}| \mathcal{S}_p) = - \sum_{i=j}^{m} \log p_\Theta(e_{i:m}| e_{1:i}).
    \label{autocompletion}
\end{equation}
Here, $m$ is the total number of entities in the complete sketch $\mathcal{S}$, and $\mathcal{S}_{p}=(e_1, ..., e_i)$ is a partial sequence containing 20\% to 80\% of the primitive entities from $\mathcal{S}$. Our model's goal is to optimize the log-likelihood of completing the sketch $\mathcal{S}$ based on these partial sequences.

We employ open-source natural language pre-trained Transformer models for our generative model, with a particular emphasis on T5 models~\cite{raffel2020exploring}. Additionally, we explore finetuning with GPT-3.5~\cite{brown2020language}. Our results indicate significant cross-domain knowledge transfer capabilities in these pre-trained models, showcasing their effectiveness in the context of CAD design.

\textbf{CAD Autoconstrain Task.}
The CAD autoconstrain task aims to autoregressively generate a set of constraints based on given geometry primitives. We model the constraint model as 
\begin{equation}
\setlength{\abovedisplayskip}{2pt}
\setlength{\belowdisplayskip}{2pt}
    \mathcal{L}(\phi) = - \log p_\Phi(\mathcal{C}| \mathcal{S}) = - \sum_{i=1}^{n} \log  p_\Phi(C_{j:n}| C_{1:j},S),
    \label{autoconstraint}
\end{equation}
where $n$ is the number of constraints, and each $\mathcal{C}_i$ is represented by a tuple of tokens indicating its constraint type and corresponding primitive entity parameters. Like autocompletion task, the autoconstrain model is trained to maximize the log-likelihood of generated tokens with respect to the ground truth. 


Aligned with \cite{seff2021vitruvion}, we focus on 13 constraint types, which are coincident, parallel, and so on. Further details about these constraints are provided in the Supplementary.

\textbf{Evaluation Metrics.} To quantitatively assess CAD generation models, we introduce three CAD-specific metrics: \textit{Sketch Accuracy}, \textit{Entity Accuracy}, and \textit{CAD F1}. 

\textit{Sketch Accuracy} measures the probability of correctly generating the entire remaining sketch in the test set.
\textit{Entity Accuracy} evaluates the likelihood of correctly generating at least one entity in the ground truth. They are calculated as:
\begin{equation}
    \textit{Sketch Accuracy} = \frac{N_s}{N}, \quad \textit{Entity Accuracy} = \frac{N_e}{N},
\end{equation}

where $N_s$ is the number of correctly generated sketches in the test set, $N_e$ is the number of generated sketches that match at least one entity with the ground truth sketch, and $N$ is the total number of sketches in the test set.
\textit{CAD F1} is a harmonic mean of precision and recall for each sketch, averaged over the dataset:
\begin{equation}
\small
    \textit{CAD F1} = \frac{1}{N} \sum^{N}_{i=1} \frac{2 \times \text{sketch precision} \times \text{sketch recall}}{(\text{sketch precision} + \text{sketch recall})},
\end{equation}
where $\textit{sketch precision} = \frac{n_c}{n_p}$, $\textit{sketch recall} = \frac{n_c}{m}$, $n_c$ is the count of correct entities in the completed sketch, $n_p$ is the total number of entities predicted, and $m$ is the number of entities in the original sketch (ground truth).

\section{CadVLM: A Vision Language Model for CAD Generation}
\label{sec:method}
In this section, we delve into the specifics of our CAD Vision Language Model (CadVLM), employing the CAD autocompletion task as an illustrative example. In our approach, the sequence of sketch primitives $\mathcal{S}$ is interpreted akin to language text, while the corresponding visual representation of the sketch is treated as image $I$. It should be noted that the text-grounded image decoder is not utilized in the CAD auto-constraint task.

\subsection{Network Framework}

CadVLM is structured as an asymmetrical encoder-decoder architecture, featuring a two-stream encoder, a text decoder based on a language model, and a lightweight image decoder for constructing the autocompleted image. This design is depicted in Figure~\ref{fig:architecture}. The two-stream encoder is tailored to process dual-modal inputs: a vision sub-encoder for the sketch image $I$ and a text sub-encoder for the prefix sequence of sketch primitives $S_p$. For text decoding, we employ a causal language transformer model, which autoregressively generates the output, i.e., the autocompleted sketch. Finally, the image decoder utilizes the latent representation tokens to construct the visual representation of the autocompleted sketch.

\begin{wrapfigure}{l}{0.5\textwidth}
    \centering
    \includegraphics[scale=0.53]{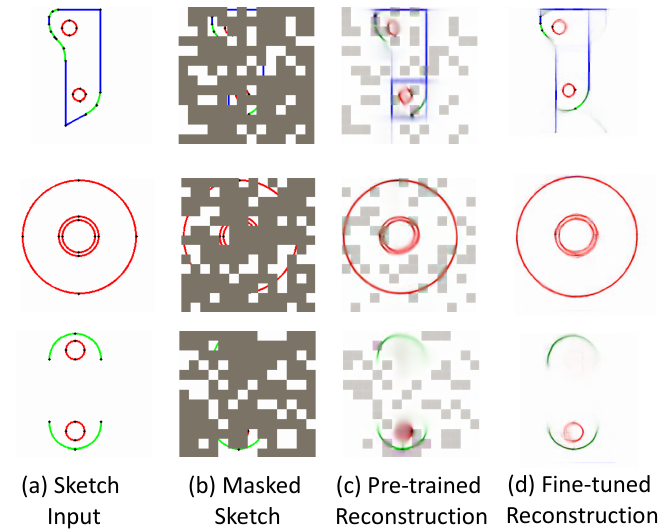}
    \caption{CAD image reconstruction results by ViTMAE. (a) Input sketches, (b) masked sketches with the mask ratio of 75\%, reconstructed results using a (c) pre-trained and (d) further fine-tuned ViT-MAE.}
    \label{fig:vitmae}
\end{wrapfigure}

\textbf{Vision Encoding Stream.}
To encode partial sketch images and harness the capabilities of an existing large vision autoencoder model, we employ a pre-trained vision transformer (ViT)\cite{he2022masked} for extracting vision features. To bridge the gap between CAD images and natural images typically processed by ViT, we fine-tune the model on the SketchGraphs dataset. This fine-tuning involves a one-epoch image reconstruction task, with a 75\% masking ratio, specifically tailored for CAD sketch data. Figure~\ref{fig:vitmae} illustrates the image reconstruction outcomes using both the original pre-trained ViT-MAE and the version fine-tuned on the SketchGraphs dataset. After finetuning, ViT-MAE shows enhanced representational performance for CAD sketches.

However, it's important to note that while some reconstructed sketches are notably accurate, they do not fully meet the requirements for parametric CAD engineering due to the vagueness and lack of parameterization in the entities. To bring image and text features to the same embedding space, we integrate a down-sampling layer and a lightweight projection module atop the vision encoder.

\textbf{Text Encoding Stream.} 
Treating the sketch primitive sequence as textual data, we first tokenize this sequence to derive a series of text tokens $T={t_1,t_2,...,t_l}$. These tokens are then processed using the encoder of a pre-trained language model, effectively encoding the sketch information in a textual format.

\textbf{Text-grounded Vision Decoding Stream.} This stream is dedicated to constructing the autocompleted pixels, mapping the fused image and text representation to the pixel-level sketch image.  The input for the vision decoder comprises the fused image and text embeddings, augmented with adaptively masked token patches to keep the same patch size as the input image. This construction process ensures that the vision embedding is grounded by textual information and also enhances the text representation.
Note that for CAD autoconstraint task, we don't reconstruct the image since the input sketches are complete sketch images already.

\textbf{Image-grounded Text Decoding Stream.} In alignment with Equations~\ref{autocompletion} and \ref{autoconstraint}, we conceptualize autocompletion as a conditional text generation process. Within this framework, the combined image and text embeddings serve as inputs to the decoder of a language model. This decoder is capable of autoregressively generating the required output. Notably, the parameters of this language model are initialized from a pre-trained model, which itself has undergone extensive pre-training on vast datasets of code~\cite{wang2023codet5+}, enhancing its effectiveness in this context.

\subsection{Training Objectives}
As shown in Figure~\ref{fig:architecture},  we jointly optimize three distinct objectives during training: one focused on understanding and two on generation.

\textbf{Image-Text Contrastive Loss (ITC)} is employed to activate the unimodal encoders, following the approach in~\cite{li2022blip}. This loss function aligns the visual and textual feature spaces by ensuring that positive image-text pairs have similar representations while contrasting with negative pairs. Such a contrastive approach has been instrumental in enhancing multimodal model training~\cite{radford2021learning, li2021align}, not only improving the alignment of visual and language features but also accelerating the convergence during training.
The image-text contrastive loss is defined as the cross-entropy between the similarity $p$ of image(I) text(T) embedding pairs and ground-truth one-hot similarity $y$ (array from [0,N], N is the number of samples), formulated as: 
\begin{equation}
    p = \frac{\text{exp}(s(I, T))}{\sum\limits_{i,j}^{N} \text{exp} (s(I_{i}, T_{j}))},  \mathcal{L}_{ITC} = \text{Cross-Entropy}(y,p).
\end{equation}

\textbf{Image Decoding Loss (IDL)} is responsible for constructing autocompletion sketch images. This process involves translating the fused visual and textual information into pixel values for each patch. IDL employs a mean squared error (MSE) loss, similarly to the approach in ViTMAE~\cite{he2022masked}, comparing the autocompleted sketch images with their ground-truth counterparts at the pixel level. The Image decoding loss can be formulated as:
\begin{equation}
    \mathcal{L}_{IDL} = \frac{1}{n} \sum_{i=1}^{n} (Y_i - \hat{Y}_i)^2,
\end{equation}
where $n$ is the number of total pixels, $Y_i$ is the ground-truth pixel value of $i$-th pixel, and $\hat{Y}_i$ is the output pixel value correspondingly. 
For CAD autoconstraint tasks, since the input is a full sketch image, we ignore IDL during the training of CAD autoconstraint models.

\textbf{Language Modeling Loss (LM)} is designed to generate text-like autocompletion sketches, focusing on constraint sequences and utilizing the image-grounded text decoder. It optimizes a cross-entropy loss to maximize the likelihood described in Equation~\ref{autocompletion}. LM endows the model with the capability to transform fused multimodal information into coherent autocompletion sketch sequences. For the CAD autoconstraint task, we use Equation~\ref{autoconstraint} as the language modeling loss. 

Therefore, the full training objective of CadVLM is:

\begin{equation}
\setlength{\abovedisplayskip}{2pt}
\setlength{\belowdisplayskip}{2pt}
    \mathcal{L} = \mathcal{L}_{ITC} +\mathcal{L}_{LM} +\mathcal{L}_{IDL}.   
\end{equation}

During inference, given incomplete sketch image and primitive pairs, we only evaluate the generated text and use that for further mechanical applications.

\section{Experiments}
We conduct extensive experiments and ablation studies on CadVLM for the CAD autocompletion and autoconstraint tasks. Quantitative assessment is conducted by measuring Entity Accuracy, Sketch Accuracy, and CAD F1. 

\subsection{Experimental settings}

\textbf{Dataset} We use the SketchGraphs~\cite{seff2020sketchgraphs} dataset for all our experiments, consisting of 15 million CAD sketches collected from Onshape\footnote{https://www.onshape.com}. Referring to~\cite{seff2021vitruvion}, we use the filtered version and further deduplicate the dataset. After deduplication, the training, validation, and test set have 626,236, 22,031, 21,979 unique sketches correspondingly. The filtered sketches are restricted to sketches comprised of arcs, circles, and lines.
To ensure a fair comparison, we normalize each sketch as in ~\cite{seff2021vitruvion}; specifically the quantized range of numerical tokens in the primitive sequence is $[1,64]$.


\textbf{Baselines} We evaluate several ablation and conditional variants of CadVLM, as well as three SOTA CAD generative model baselines: \textbf{Vitruvion}~\cite{seff2021vitruvion}, \quad \textbf{Deepcad}~\cite{wu2021deepcad} and \textbf{SketchGraphs}~\cite{seff2020sketchgraphs}.
Since SketchGraphs does not incorporate continuous primitive parameters such as coordinate value, we only compare with the CAD autoconstraint task. And DeepCAD only considers parametrized command sequences, which are incompatible with CAD constraints. We primarily evaluate its performance on CAD autocompletion tasks.


\textbf{Settings} For the image encoder and decoder, we choose the encoder and decoder of state-of-the-art pre-trained ViT-MAE-base from~\cite{he2022masked}. Following~\cite{he2022masked} we set the patch size and stride both as 32. We render all sketch images with size of $224\times224\times3$.
For the text encoder and text decoder, we use the encoder and decoder of pre-trained CodeT5+~\cite{wang2023codet5+} correspondingly, which is an instruction-tuned code LLM with 770M parameters trained on various code datasets.
We train the autocompletion model and autoconstraint model both for 30 epochs, using the Adam optimizer with decoupled weight decay  regularization~\cite{loshchilov2017decoupled}, with the learning rate set according to the  CosineAnnealingLR~\cite{loshchilov2016sgdr} scheduler. The initial learning rate is set to $3e-4$, with batch size 32. All our experiments are run using 8 NVIDIA A100 GPUs, and our models take between 30-40 hours to train.
In the CAD Autocompletion experiments, we randomly sample 20\%-80\% of sketch entities as input and aim to predict the remaining entities from the ground truth sketch.

\begin{table}[t]
\centering
\caption{Quantitative sketch generative model comparison results for CAD autocompletion task. Ske-Acc means the probability of generating the correct full remaining sketch and Ent-Acc means the probability of generating at least one entity of the ground truth. \textbf{Bold} shows the best results of each metric. For all metrics the higher the better.}
\begin{tabular}{cccc}
\toprule
\multicolumn{1}{c|}{Model} & \multicolumn{1}{c}{Ske-Acc} & \multicolumn{1}{c}{Ent-Acc} & CAD-F1 \\ \hline
\multicolumn{1}{c|}{Vitruvion}          & 3.0\%  & 40.7\% & 11.3\% \\ \hline
\multicolumn{1}{c|}{Deepcad}          & 1.2\%  & 23.1\% & 8.9\% \\ \hline
\multicolumn{1}{c|}{ChatGPT-15\%}       & 6.8\%  & 41.3\% & 21.2\% \\
\multicolumn{1}{c|}{ChatGPT-7\%}        & 7.6\%  & 43.8\% & 22.2\% \\ 
\multicolumn{1}{c|}{ChatGPT-1\%}        & 6.5\%  & 43.7\% & 22.5\% \\ \hline
\multicolumn{1}{c|}{CadVLM-Text}        & 21.2\%  & 66.7\% & 42.8\% \\ 
\multicolumn{1}{c|}{CadVLM-w/o-IDL}     & 22.6\% & \textbf{68.4}\% & 44.2\%\\ 
\multicolumn{1}{c|}{CadVLM-w/o-ITC}     & 22.8\% & 67.6\% & 43.9\% \\ 
\multicolumn{1}{c|}{CadVLM-w/o-IDL\&ITC} & 23.0\% & 67.7\% & 44.1\% \\ \hline
\multicolumn{1}{c|}{\textbf{CadVLM}}    & \textbf{23.8\%} & 68.3\% & \textbf{45.2\%} \\ \hline
\end{tabular}
\label{tab:autocompletion}
\end{table}

\begin{table}[t]
\centering
\caption{Quantitative sketch generative model comparison results for CAD autoconstraint task. For all metrics higher is better.}
\begin{tabular}{cccc}
\toprule
\multicolumn{1}{c|}{Model} & \multicolumn{1}{c}{Ske-Acc} & \multicolumn{1}{c}{Ent-Acc} & CAD-F1 \\ \hline
\multicolumn{1}{c|}{SketchGraphs}       & 0.48\% & 93.0\% & 42.0\% \\ \hline
\multicolumn{1}{c|}{Vitruvion}          & 10.7\% & 94.6\% & 71.6\% \\ \hline
\multicolumn{1}{c|}{CadVLM-Text}        & 11.5\% & 95.0\% & 72.6\% \\ 
\multicolumn{1}{c|}{CadVLM-w/o-ITC} & \textbf{12.1\%} & \textbf{95.4\%} & \textbf{74.5\%} \\ \hline
\multicolumn{1}{c|}{\textbf{CadVLM}}    & 11.3\% & 94.9\% & 73.8\% \\ 
\toprule
\end{tabular}
\label{tab:autoconstraint}
\end{table}

\begin{figure*}[h]
    \centering
    \includegraphics[scale=0.36]{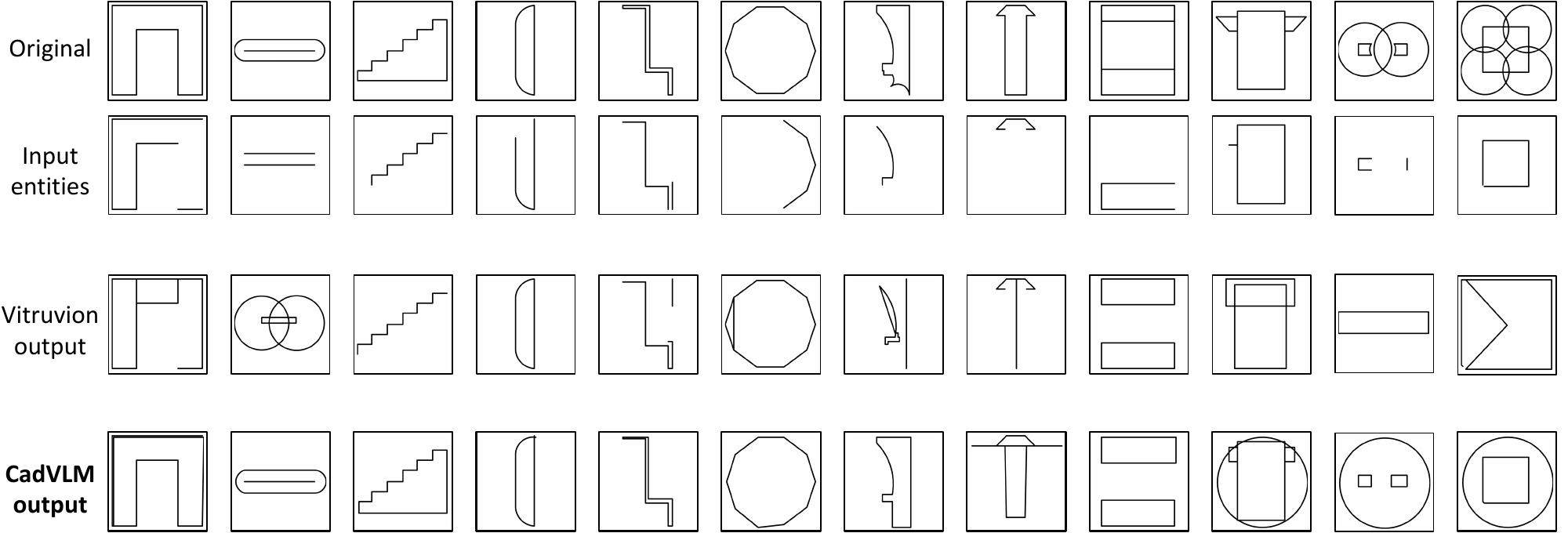}
    \caption{Comparative Analysis of Autocompletion in CAD Design. Top row: Random samples from the SketchGraphs test dataset. Second row: Initial input entities serving as the primer for the autocompletion task. Third and fourth row: Autocompletion results produced by Vitruvion and CadVLM.} 
    \label{fig:samples}
\end{figure*}

\subsection{Quantitative comparison with baseline sketch generative models}
Table~\ref{tab:autocompletion} reports results on the \textbf{CAD Autocompletion} task. We find that Vitruvion achieves 3\% on Sketch Accuracy and 40.7\% on Entity Accuracy, meaning it can accurately predict 3\% of sketches and correctly identifies at least one entity in 40.7\% of sketches. Deepcad performs worst since it only have 4 layers of encoder and decoder, makes it hard to represent complex CAD sketches. Our CadVLM outperforms Virtruvion on Sketch Accuracy, Entity Accuracy, and CAD-F1 by 20.8\%, 28.3\%, and 39.9\% respectively. These results demonstrate the effectiveness of pre-trained foundational models and image modality. A closer examination of the qualitative results in Figure~\ref{fig:samples} reveals that Vitruvion produces many invalid results missing symmetric curves or lines. Sketches from CadVLM are better in terms of quality with more complex shapes and stronger symmetry. 


\textbf{ChatGPT Series Model Finetune} To evaluate the ability of a general LLM on the CAD autocompletion task, we fine-tuned ChatGPT on three subsets of 15\%, 7\%, and 1\% of the SketchGraphs dataset. The results indicate that the ChatGPT-series outperforms Vitruvion on all three metrics, especially on Sketch Accuracy and CAD-F1. However, there were no significant performance improvements when finetuning on larger portions of the training dataset. One possible reason is we only finetune ChatGPT for less than 5 epochs due to training costs, making it hard for ChatGPT to adapt to CAD data.

\textbf{CadVLM variants} To better understand the source of performance gains, we conducted a comparative analysis of different variants of our proposed CadVLM: CadVLM-text,  CadVLM-w/o-ITC, 
CadVLM-w/o-IDL\&ITC, and CadVLM-w/o-IDL. CadVLM-text relies solely on the sketch primitive sequence text as its input, optimizing for language modeling loss. In contrast, CadVLM-w/o-IDL utilizes ITC and LM loss for its training objectives, CadVLM-w/o-ITC focuses on IDL and LM, and CadVLM-w/o-IDL\&ITC prioritizes only the LM objective. The findings reveal that CadVLM-Text underperforms compared to all other CadVLM variants across all three metrics, which underlines the value of the sketch image modality in providing diverse geometric information alongside sequence text.

Interestingly, CadVLM-w/o-IDL emerges as the superior performer among the CadVLM variants, highlighting the significance of aligning image and text embeddings in multimodal modeling. Incorporating Image Decoding Loss (IDL) further enhances the alignment between image and text embeddings, leading CadVLM to show a notable improvement in all three metrics.

For the \textbf{CAD Autoconstraint} task, as shown in Table~\ref{tab:autoconstraint}, SketchGraphs struggles to predict the whole sketch correctly, with only 0.48\% Sketch Accuracy. This suggests that relying solely on graph structures of sketches is insufficient for handling long sequences of sketch primitives. In contrast, Vitruvion achieves 10.7\% on Sketch Accuracy, showing the necessity to model primitives.
For the CAD Autoconstraint task it is less challenging to predict at least one constraint correctly but more challenging to accurately predict all constraints, easily caused by potential data labeling inconsistencies. As is well-known, minor discrepancies in primitive values can drastically alter constraints, such as changing them from parallel to coincident.

The CadVLM series model exceeds the performance of both SketchGraphs and Vitruvion across all metrics. 
CadVLM-Text surpasses CadVLM on Ske-Acc and Ent-Acc since the pretraining task of ViTMAE has a great discrepancy with CAD autoconstraint tasks. Therefore aligning the image and text representation directly might not be an optimal way for CAD autoconstraint task.
Notably, CadVLM-w/o-ITC emerges as the top performer within the CadVLM series. This may be attributed to its more substantial text decoder 
and utilization of multimodal information
, which enables the model to more effectively integrate knowledge from both primitives and images to inform geometric information.

\begin{figure*}[t]
	\begin{minipage}{0.328\linewidth}
		\centerline{\includegraphics[width=\textwidth]{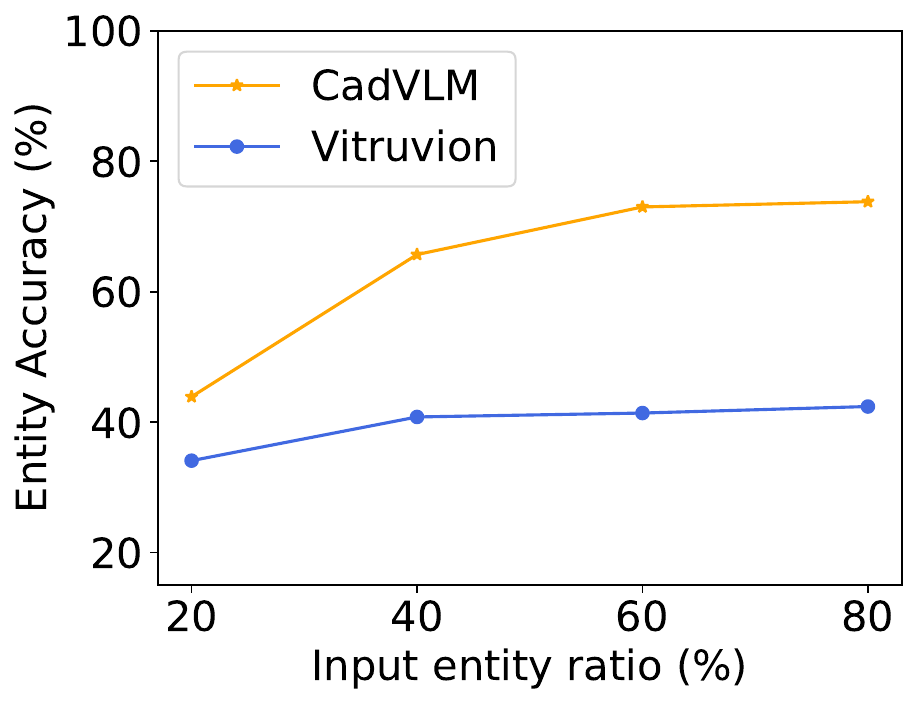}}
		\centerline{(a)}
	\end{minipage}
	\begin{minipage}{0.328\linewidth}
		\centerline{\includegraphics[width=\textwidth]{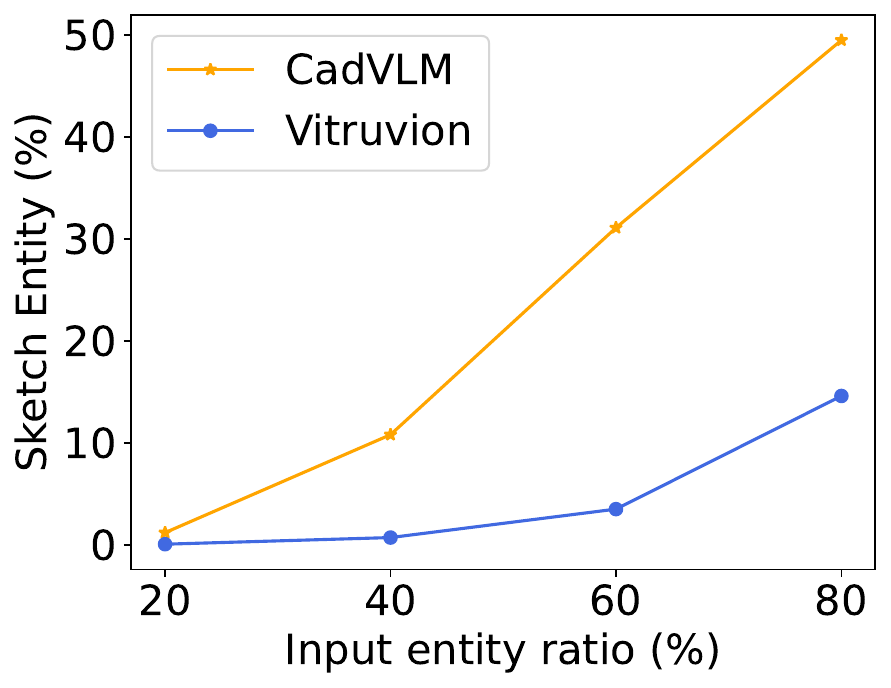}}
		\centerline{(b)}
	\end{minipage}
	\begin{minipage}{0.328\linewidth}
		\centerline{\includegraphics[width=\textwidth]{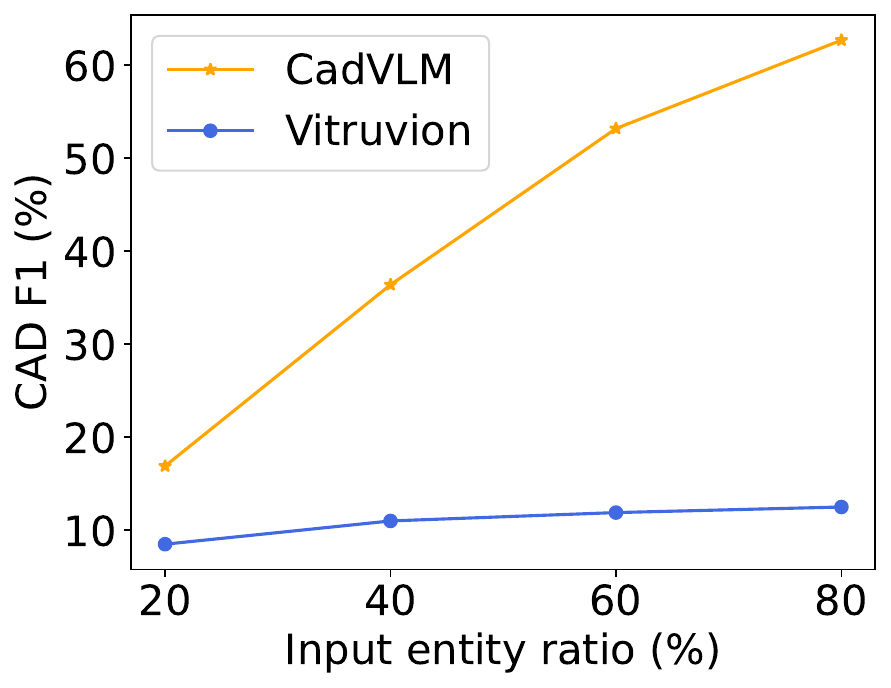}}
		\centerline{(c)}
	\end{minipage}
	\caption{Effect of input entity ratio to (a) Entity Accuracy, (b) Sketch Accuracy, and (c) CAD F1.}
	\label{fig:ratiovsacc}
\end{figure*}

\begin{minipage}[t]{0.53\textwidth}
\makeatletter\def\@captype{table}
\small
\centering
\caption{Quantitative results for image conditioned primitive generation. \textbf{Bold} represents the best results of all metrics.}
\begin{tabular}{c|c|c|c}
\toprule
\begin{tabular}[c]{@{}c@{}}Conditioned\\ -Image\end{tabular}                  & Model     & Ent-Acc         & CAD-F1          \\ \hline
\multirow{2}{*}{Precise}          & Vitruvion & 83.8\%          & 35.8\%          \\ \cline{2-4} 
                                  & CadVLM    & \textbf{98.0\%} & \textbf{62.7\%} \\ \hline
\multirow{2}{*}{Hand-drawn}       & Vitruvion & 84.2\%          & 38.4\%          \\ \cline{2-4} 
                                  & CadVLM    & \textbf{99.4\%} & \textbf{60.1\%} \\ \hline
\multirow{2}{*}{\begin{tabular}[c]{@{}c@{}}Noisy\\ hand-drawn\end{tabular}} & Vitruvion & 83.2\%          & 35.8\%          \\ \cline{2-4} 
                                  & CadVLM    & \textbf{94.4\%} & \textbf{56.5\%} \\ \toprule
\end{tabular}
\label{tab:image_cond}
\end{minipage}
\begin{minipage}[t]{0.42\textwidth}
\makeatletter\def\@captype{table}
\caption{Effect of adding the inductive bias of entity-level embeddings to the sketch sequence.}
\centering
\small
\begin{tabular}{c|cc}
\toprule
Model   & \begin{tabular}[c]{@{}c@{}}CadVLM\\ -Text\end{tabular} & \begin{tabular}[c]{@{}c@{}}w/-Inductive \\ Bias\end{tabular}  \\
\midrule
Ent-Acc & 66.7\%      & 67.5\%             \\
\midrule
Ske-Acc & 21.2\%      & 22.2\%            \\
\midrule
CAD-F1  & 42.8\%      & 43.7\%            \\
\toprule
\end{tabular}
\label{tab:inductive_bias}
\end{minipage}

\subsection{Ablation Study}
We conducted several ablation studies to assess the performance of CadVLM on image-conditioned primitive generation tasks and investigate the effectiveness of each component within CadVLM.

\textbf{Image Conditional Generation.} 
Following~\cite{seff2021vitruvion}, we conduct a series of experiments to evaluate the effectiveness of CadVLM modeling primitive sequences conditioned on the images of sketches. As designers may begin their designs with hand-drawn sketches before transitioning to CAD programs, accurately inferring parametric CAD from images or scans can dramatically reduce the effort of translating from rough paper sketches to the CAD model.

When conditioning only on sketch images, the text encoder of CadVLM is ignored. We only use the image encoder, image decoder, and text decoder. The overall model is trained to maximize the probability of the ground truth primitives portrayed in the image, conditioned on a sequence of patch embeddings.

The results of Table~\ref{tab:image_cond} show three kinds of sketch image rendering input to imitate different settings in the real world: precise renderings, rendering from the hand-drawn simulator of~\cite{seff2021vitruvion}, and renderings with random affine augmentations of the hand-drawn simulator. As both Vitruvion and CadVLM can seldom predict the sketch exactly correctly, we ignore Sketch Accuracy in our image conditional generation experiments. The results show that CadVLM outperforms Vitruvion on both Entity Accuracy and CAD F1. Vitruvion performs best on hand-drawn images with 83.8\% Entity Accuracy and 38.4\% CAD F1. While CadVLM achieves almost 99.4\% Entity Accuracy and 62.7\% CAD F1.
Conditioning on precise rendered images acheives the best performance.


\textbf{Entity Level Modelling.}
As described earlier, each CAD sketch is comprised of multiple entities. We considered adding this inductive bias to the model by processing each entity separately and then combining their embeddings with the initial sequence of all sketch tokens. For this purpose, we first prepend a special token \textlangle ENTITY\textrangle~to each of the entity input sequences. Subsequently, the entities from all the sketches in the batch are fed into the model's encoder in parallel.
Next, the embeddings of each entity are gathered for each sketch. For each sketch, we concatenate these embeddings with the original sequence tokens using another special token \textlangle TOKEN\textrangle~as a delimiter for the second part of the sequence. The decoder remains unchanged. As shown in Table \ref{tab:inductive_bias}, incorporating this inductive bias increases computation and memory demands while only marginally enhancing the performance.




\textbf{Impact of different input entity ratios for CAD autocompletion.} To explore the impact of different input entity ratios for CAD autocompletion, we further test CAD autocompletion results with different input entity ratios: 20\%, 40\%, 60\%, and 80\%. Compared with the experiments in Sec 5.4, we specify the sampled ratio of input entities. As shown in Figure~\ref{fig:ratiovsacc}, CadVLM outperforms Vitruvion on all input entity ratios. In particular, CadVLM achieves 73.8\% Entity Accuracy, 49.5\% Sketch Accuracy, and 62.7\% CAD F1 when given 80\% of entities as input. For the hardest setting when given only 20\% of entities as input, CadVLM can still achieve 43.9\% Entity Accuracy and 16.9\% CAD-F1, which proves that CadVLM has great potential to be used in real world settings.

\begin{table}[t]
\centering
\caption{Effect of using different architectures for CadVLM-Text. CodeT5+ 770M is the encoder-decoder for text we use in our main experiments. CodeT5+* is the result of training from scratch rather than pre-trained. }
\begin{tabular}{c|cccc}
\toprule
Model   & \#Parameter                  & Ent-Acc     & Ske-Acc        & CAD-F1          \\ 
\hline

CodeT5+   & 220M &  66.5\%            & 21.5\%               &   42.9\% \\
$\text{CodeT5+}^*$ & 220M & 65.0\%             & 18.9\%             & 40.3\% \\ 

CodeT5+   & 770M &  66.7\%            & 21.2\%              &   42.8\% \\
ByT5-base & 580M & 66.9\%                  & 22.0\%                    & 43.1\%          \\ 

ByT5-large & 1.2B & 66.8\%                  & 20.6\%                    & 41.9\%          \\

T5-base   & 223M & 66.2\%             & 20.2\%             & 41.7\% \\ 

T5-large   & 783M & 66.8\%             & 21.2\%             & 42.6\% \\ 

\toprule
\end{tabular}
\label{tab: different language models}
\end{table}
\textbf{Effect of using different language models for text encoder and decoder.} To better compare the performance of different text encoder-decoder, we experiment using different pre-trained foundation models as text encoder and decoder. To eliminate the influence of the image encoder and decoder, we only consider the CadVLM-Text variant. For the choice of text encoder and decoder, we also compare the performance of different Encoder-Decoder pre-trained language models on CAD autocompletion. The T5 family of models are among the most recognized and widely used encoder-decoder models in the open-source community. Therefore, we experimented with T5 ~\cite{raffel2020exploring}, ByT5 ~\cite{xue2022byt5} and CodeT5+ ~\cite{wang2023codet5+}. The results are shown in Table \ref{tab: different language models}. The effect of model size is clear for T5 and CodeT5+ but not ByT5. Also, The cost-benefit of solely making the model bigger becomes less favorable over time as the improvements in performance start to plateau. Another observation is that, because of the nature of the data which is closer to a code sequence than natural language, CodeT5+, which is fine-tuned on code, exhibits better performance. 

We also trained CodeT5+ from scratch to examine the impact of pre-training on natural language and code. As evidenced in Table~\ref{tab: different language models}, training from scratch falls short in all three metrics. Additionally, training from scratch converged much slower than finetuning. Despite the fact that our representation of sketches as code is relatively unique and not found in natural language, the rules and token embeddings learned from text are very helpful for domain adaptation. 

\section{Conclusion}
In this study, we introduce CadVLM, an innovative end-to-end vision-language model tailored for modeling parametric CAD sketches, which are pivotal in modern mechanical design. CadVLM harnesses the commonsense knowledge and reasoning abilities inherent in large language models, enabling it to adeptly handle the complex geometric reasoning required for CAD sketches. Our experimental results demonstrate that CadVLM outperforms existing state-of-the-art models in both autocompletion and autoconstraint tasks, highlighting the potent enhancement that pre-trained models bring to CAD construction.


\section*{Acknowledgements}
This work is supported by the Canada CIFAR AI Chair Program and the Canada NSERC Discovery Grant (RGPIN-2021-03115).

\bibliographystyle{splncs04}
\bibliography{main}

\begin{thebibliography}{10}
\providecommand{\url}[1]{\texttt{#1}}
\providecommand{\urlprefix}{URL }
\providecommand{\doi}[1]{https://doi.org/#1}

\bibitem{azerbayev2023llemma}
Azerbayev, Z., Schoelkopf, H., Paster, K., Santos, M.D., McAleer, S., Jiang, A.Q., Deng, J., Biderman, S., Welleck, S.: Llemma: An open language model for mathematics. arXiv preprint arXiv:2310.10631  (2023)

\bibitem{brown2020language}
Brown, T., Mann, B., Ryder, N., Subbiah, M., Kaplan, J.D., Dhariwal, P., Neelakantan, A., Shyam, P., Sastry, G., Askell, A., et~al.: Language models are few-shot learners. Advances in neural information processing systems  \textbf{33},  1877--1901 (2020)

\bibitem{camba2016parametric}
Camba, J.D., Contero, M., Company, P.: Parametric cad modeling: An analysis of strategies for design reusability. Computer-Aided Design  \textbf{74},  18--31 (2016)

\bibitem{chen2023medblip}
Chen, Q., Hu, X., Wang, Z., Hong, Y.: Medblip: Bootstrapping language-image pre-training from 3d medical images and texts. arXiv preprint arXiv:2305.10799  (2023)

\bibitem{chen2022pali}
Chen, X., Wang, X., Changpinyo, S., Piergiovanni, A., Padlewski, P., Salz, D., Goodman, S., Grycner, A., Mustafa, B., Beyer, L., et~al.: Pali: A jointly-scaled multilingual language-image model. arXiv preprint arXiv:2209.06794  (2022)

\bibitem{chen2024camml}
Chen, Y., Zhang, S., Han, B., He, T., Li, B.: Camml: Context-aware multimodal learner for large models. arXiv preprint arXiv:2401.03149  (2024)

\bibitem{floridi2020gpt}
Floridi, L., Chiriatti, M.: Gpt-3: Its nature, scope, limits, and consequences. Minds and Machines  \textbf{30},  681--694 (2020)

\bibitem{ganin2021computeraided}
Ganin, Y., Bartunov, S., Li, Y., Keller, E., Saliceti, S.: Computer-aided design as language (2021)

\bibitem{he2022masked}
He, K., Chen, X., Xie, S., Li, Y., Doll{\'a}r, P., Girshick, R.: Masked autoencoders are scalable vision learners. In: Proceedings of the IEEE/CVF conference on computer vision and pattern recognition. pp. 16000--16009 (2022)

\bibitem{jones2023self}
Jones, B.T., Hu, M., Kodnongbua, M., Kim, V.G., Schulz, A.: Self-supervised representation learning for cad. In: Proceedings of the IEEE/CVF Conference on Computer Vision and Pattern Recognition. pp. 21327--21336 (2023)

\bibitem{kenton2019bert}
Kenton, J.D.M.W.C., Toutanova, L.K.: Bert: Pre-training of deep bidirectional transformers for language understanding. In: Proceedings of naacL-HLT. vol.~1, p.~2 (2019)

\bibitem{li2022blip}
Li, J., Li, D., Xiong, C., Hoi, S.: Blip: Bootstrapping language-image pre-training for unified vision-language understanding and generation. In: International Conference on Machine Learning. pp. 12888--12900. PMLR (2022)

\bibitem{li2021align}
Li, J., Selvaraju, R., Gotmare, A., Joty, S., Xiong, C., Hoi, S.C.H.: Align before fuse: Vision and language representation learning with momentum distillation. Advances in neural information processing systems  \textbf{34},  9694--9705 (2021)

\bibitem{li2023secad}
Li, P., Guo, J., Zhang, X., Yan, D.M.: Secad-net: Self-supervised cad reconstruction by learning sketch-extrude operations. In: Proceedings of the IEEE/CVF Conference on Computer Vision and Pattern Recognition. pp. 16816--16826 (2023)

\bibitem{lindsay2018identifying}
Lindsay, A., Paterson, A., Graham, I.: Identifying and quantifying inefficiencies within industrial parametric cad models. In: Advances in Manufacturing Technology XXXII: Proceedings of the 16th International Conference on Manufacturing Research. vol.~8, p.~227 (2018)

\bibitem{loshchilov2016sgdr}
Loshchilov, I., Hutter, F.: Sgdr: Stochastic gradient descent with warm restarts. arXiv preprint arXiv:1608.03983  (2016)

\bibitem{loshchilov2017decoupled}
Loshchilov, I., Hutter, F.: Decoupled weight decay regularization. arXiv preprint arXiv:1711.05101  (2017)

\bibitem{luo2023wizardmath}
Luo, H., Sun, Q., Xu, C., Zhao, P., Lou, J., Tao, C., Geng, X., Lin, Q., Chen, S., Zhang, D.: Wizardmath: Empowering mathematical reasoning for large language models via reinforced evol-instruct. arXiv preprint arXiv:2308.09583  (2023)

\bibitem{para2021sketchgen}
Para, W., Bhat, S., Guerrero, P., Kelly, T., Mitra, N., Guibas, L.J., Wonka, P.: Sketchgen: Generating constrained cad sketches. Advances in Neural Information Processing Systems  \textbf{34},  5077--5088 (2021)

\bibitem{radford2021learning}
Radford, A., Kim, J.W., Hallacy, C., Ramesh, A., Goh, G., Agarwal, S., Sastry, G., Askell, A., Mishkin, P., Clark, J., et~al.: Learning transferable visual models from natural language supervision. In: International conference on machine learning. pp. 8748--8763. PMLR (2021)

\bibitem{raffel2020exploring}
Raffel, C., Shazeer, N., Roberts, A., Lee, K., Narang, S., Matena, M., Zhou, Y., Li, W., Liu, P.J.: Exploring the limits of transfer learning with a unified text-to-text transformer. The Journal of Machine Learning Research  \textbf{21}(1),  5485--5551 (2020)

\bibitem{seff2020sketchgraphs}
Seff, A., Ovadia, Y., Zhou, W., Adams, R.P.: Sketchgraphs: A large-scale dataset for modeling relational geometry in computer-aided design. arXiv preprint arXiv:2007.08506  (2020)

\bibitem{seff2021vitruvion}
Seff, A., Zhou, W., Richardson, N., Adams, R.P.: Vitruvion: A generative model of parametric cad sketches. In: International Conference on Learning Representations (2021)

\bibitem{singhal2022large}
Singhal, K., Azizi, S., Tu, T., Mahdavi, S.S., Wei, J., Chung, H.W., Scales, N., Tanwani, A., Cole-Lewis, H., Pfohl, S., et~al.: Large language models encode clinical knowledge. arXiv preprint arXiv:2212.13138  (2022)

\bibitem{singhal2023towards}
Singhal, K., Tu, T., Gottweis, J., Sayres, R., Wulczyn, E., Hou, L., Clark, K., Pfohl, S., Cole-Lewis, H., Neal, D., et~al.: Towards expert-level medical question answering with large language models. arXiv preprint arXiv:2305.09617  (2023)

\bibitem{vaswani2017attention}
Vaswani, A., Shazeer, N., Parmar, N., Uszkoreit, J., Jones, L., Gomez, A.N., Kaiser, {\L}., Polosukhin, I.: Attention is all you need. Advances in neural information processing systems  \textbf{30} (2017)

\bibitem{wang2023codet5+}
Wang, Y., Le, H., Gotmare, A.D., Bui, N.D., Li, J., Hoi, S.C.: Codet5+: Open code large language models for code understanding and generation. arXiv preprint arXiv:2305.07922  (2023)

\bibitem{willis2021engineering}
Willis, K.D., Jayaraman, P.K., Lambourne, J.G., Chu, H., Pu, Y.: Engineering sketch generation for computer-aided design. In: Proceedings of the IEEE/CVF Conference on Computer Vision and Pattern Recognition. pp. 2105--2114 (2021)

\bibitem{wu2021deepcad}
Wu, R., Xiao, C., Zheng, C.: Deepcad: A deep generative network for computer-aided design models. In: Proceedings of the IEEE/CVF International Conference on Computer Vision. pp. 6772--6782 (2021)

\bibitem{xu2023hierarchical}
Xu, X., Jayaraman, P.K., Lambourne, J.G., Willis, K.D., Furukawa, Y.: Hierarchical neural coding for controllable cad model generation. arXiv preprint arXiv:2307.00149  (2023)

\bibitem{xu2022skexgen}
Xu, X., Willis, K.D., Lambourne, J.G., Cheng, C.Y., Jayaraman, P.K., Furukawa, Y.: Skexgen: Autoregressive generation of cad construction sequences with disentangled codebooks. In: International Conference on Machine Learning. pp. 24698--24724. PMLR (2022)

\bibitem{xue2022byt5}
Xue, L., Barua, A., Constant, N., Al-Rfou, R., Narang, S., Kale, M., Roberts, A., Raffel, C.: Byt5: Towards a token-free future with pre-trained byte-to-byte models. Transactions of the Association for Computational Linguistics  \textbf{10},  291--306 (2022)

\bibitem{zha2023tablegpt}
Zha, L., Zhou, J., Li, L., Wang, R., Huang, Q., Yang, S., Yuan, J., Su, C., Li, X., Su, A., et~al.: Tablegpt: Towards unifying tables, nature language and commands into one gpt. arXiv preprint arXiv:2307.08674  (2023)

\end{thebibliography}

\appendix

\newpage
\section{Sketch Parameters}
As mentioned in the main text, all primitives are represented as numerical tokens, and all constraints are constructed by at least one primitive, indicated by the corresponding edge's primitive tokens, followed by the constraint type token.
\begin{figure*}[!h]
    \centering
    \includegraphics[scale=0.3]{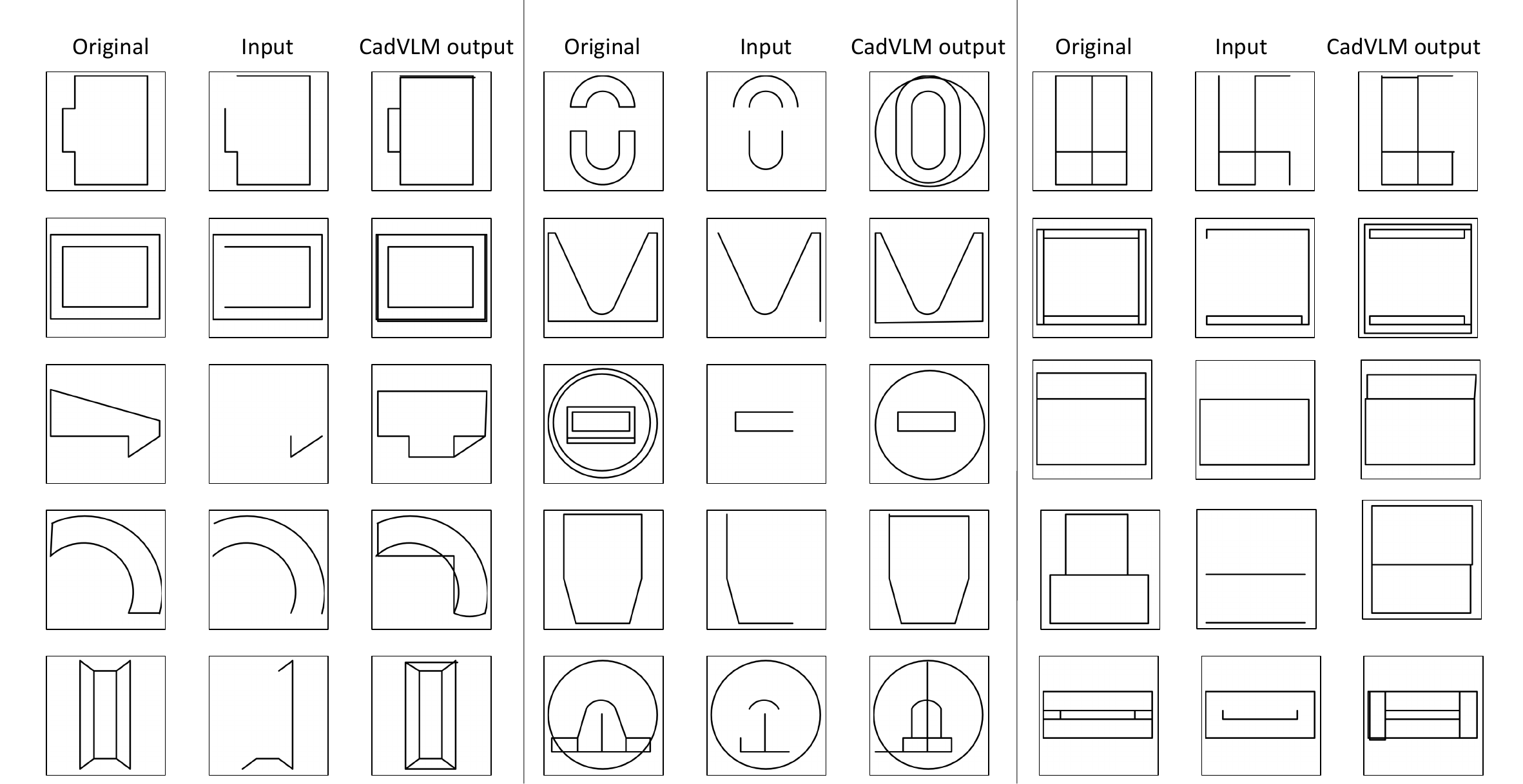}
    \caption{More CAD autocompletion Results.}
    \label{fig:supp_samples}
\end{figure*}

\begin{figure*}
    \centering
    \includegraphics[scale=0.33]{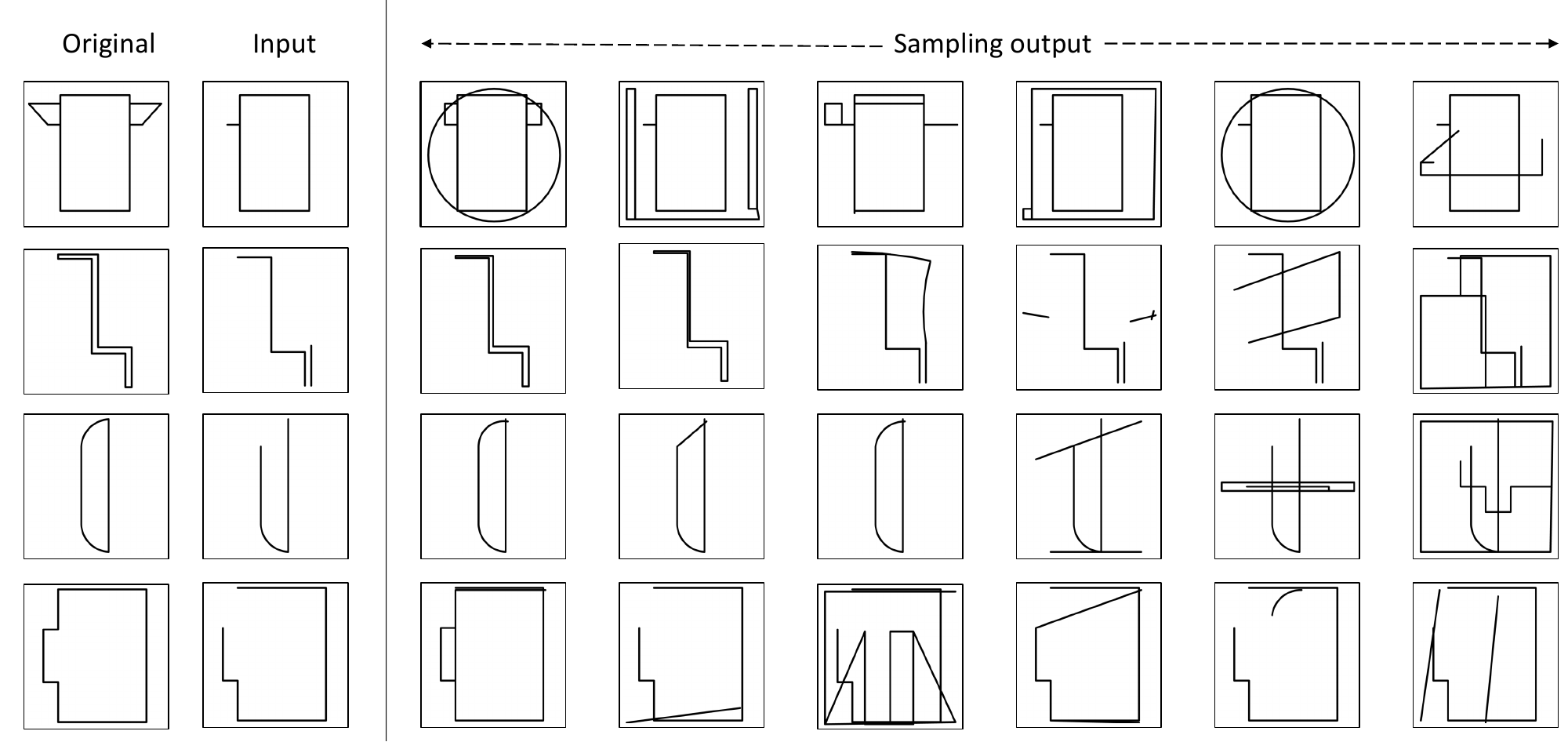}
    \caption{CAD Autocompletion Multisampling Results.}
    \label{fig:multiple_sampling}
\end{figure*}

For constraints in the sketch, we considered coincident, concentric, equal, fix, horizontal, midpoint, normal, offset, parallel, perpendicular, quadrant, tangent, and vertical. Each constraint type is represented as a pre-specified value token, as described in Table~\ref{tab:cons}.

\begin{table}[h]
    \centering
    \caption{Mapping for constraint types to the reserved set of value tokens.}
\begin{tabular}{|c|c|}
\hline
Constraint Type & Value Token \\ \hline
Coincident      & 65          \\ \hline
Concentric      & 66          \\ \hline
Equal           & 67          \\ \hline
Fix             & 68          \\ \hline
Horizontal      & 69          \\ \hline
Midpoint        & 70          \\ \hline
Normal          & 71          \\ \hline
Offset          & 72          \\ \hline
Parallel        & 73          \\ \hline
Perpendicular   & 74          \\ \hline
Quadrant        & 75          \\ \hline
Tangent         & 76          \\ \hline
Vertical        & 77          \\ \hline
\end{tabular}
    \label{tab:cons}
\end{table}

\section{Additional Evaluation}
As shown in Figure~\ref{fig:supp_samples} we provide more CAD generative model output samples.

\subsection{Multiple sampling outputs}
As shown in Figure~\ref{fig:multiple_sampling}, CadVLM can also generate multiple outputs using nucleus sampling with the cumulative probability parameter of $p=0.9$.


\section{Experimental Details}
This section provides further details of the model architecture and training regime.

\textbf{Model Architecture.} Our CadVLM all share a main transformer-based pre-trained language model as text encoder and decoder, which is responsible for processing the sequence of primitives or constraints. And a Vision transformer as image encoder and decoder for processing the sketch image. The text and image transformers architecture are identical across all the models. The text encoder and decoder are both composed of 24 layers transformer with 16 attention heads of each layer, and a total embedding dimension of 1024. No significant hyperparameter optimization was performed. The image encoder is composed of 12 layers transformer with 16 attention heads and 768 hidden size.

Additionally, we use GPT-3.5-Turbo for the ChatGPT series fine-tuning experiments. The loss is the default cross-entropy loss for language models: $\mathcal{L}_{LM}$ in Eq.(7). We use the tokenizer from the CodeT5+ model. The total number of parameters for our CadVLM model is 854M (using CodeT5+ 770M as text encoder and text decoder).



Especially for image-conditioned generation task, as the sketch primitives are unknown, we only input image modality to CadVLM. Therefore text encoder is ignored in this setting.
\end{document}